\documentclass[10pt,twocolumn,letterpaper]{article}

\usepackage{iccv}              

\usepackage[pagebackref,breaklinks,colorlinks,allcolors=iccvblue]{hyperref}
\usepackage{graphicx}
\usepackage{amsmath}
\usepackage{amssymb}
\usepackage{booktabs}
\usepackage{multirow}
\usepackage{bbm}
\usepackage{amsmath}
\usepackage{enumitem}

\usepackage[usenames,dvipsnames]{xcolor}
\definecolor{iccvblue}{rgb}{0.21,0.49,0.74}
\usepackage{pifont}
\usepackage{graphicx}
\usepackage{colortbl}
\usepackage{arydshln}
\usepackage{tikz}

\usepackage{makecell}

\usepackage{algorithm,algcompatible} 
\algnewcommand\INPUT{\item[\textbf{Input:}]}
\algnewcommand\OUTPUT{\item[\textbf{Output:}]}

\usepackage[accsupp]{axessibility}

\definecolor{darkgreen}{rgb}{0.2, 0.7, 0.1}

\definecolor{Gray}{gray}{0.8}
\definecolor{LG}{gray}{.92}

\newcommand*\colourcheck[1]{%
  \expandafter\newcommand\csname #1check\endcsname{\textcolor{#1}{\ding{51}}}%
}
\colourcheck{green}
\colourcheck{darkgreen}

\usepackage{multirow}
\usepackage{adjustbox}
\usepackage{array}
\usepackage{wrapfig,lipsum,booktabs}
\usepackage{amsmath,amssymb}
\usepackage{enumitem}
\usetikzlibrary{decorations.pathmorphing}
\definecolor{darkgreen}{rgb}{0.2, 0.7, 0.1}
\newcommand*\colourxmark[1]{%
  \expandafter\newcommand\csname #1xmark\endcsname{\textcolor{#1}{\ding{55}}}%
}
\colourxmark{red}
\usepackage{arydshln}

\usepackage{stmaryrd}
\usepackage{trimclip}

\makeatletter
\DeclareRobustCommand{\shortto}{%
  \mathrel{\mathpalette\short@to\relax}%
}

\newcommand{\short@to}[2]{%
  \clipbox{{.3\width} 0 0 0}{$\m@th#1\vphantom{+}{\shortrightarrow}$}%
  }
\makeatother

\def\paperID{4937} 
\def\confName{ICCV}
\def\confYear{2025}

\title{Unleashing the Temporal Potential of Stereo Event Cameras for Continuous-Time 3D Object Detection}

\author{Jae-Young Kang$^{*}$, Hoonhee Cho$^{*}$, and Kuk-Jin Yoon  \\
KAIST\\
{\tt\small \{mickeykang,gnsgnsgml,kjyoon\}@kaist.ac.kr}
}

\begin{document}

\maketitle
\begin{abstract}
3D object detection is essential for autonomous systems, enabling precise localization and dimension estimation. While LiDAR and RGB cameras are widely used, their fixed frame rates create perception gaps in high-speed scenarios. Event cameras, with their asynchronous nature and high temporal resolution, offer a solution by capturing motion continuously. The recent approach, which integrates event cameras with conventional sensors for continuous-time detection, struggles in fast-motion scenarios due to its dependency on synchronized sensors. We propose a novel stereo 3D object detection framework that relies solely on event cameras, eliminating the need for conventional 3D sensors. To compensate for the lack of semantic and geometric information in event data, we introduce a dual filter mechanism that extracts both. Additionally, we enhance regression by aligning bounding boxes with object-centric information. Experiments show that our method outperforms prior approaches in dynamic environments, demonstrating the potential of event cameras for robust, continuous-time 3D perception. The code is available at \url{https://github.com/mickeykang16/Ev-Stereo3D}.
\end{abstract}
 
\def\thefootnote{*}\footnotetext{Equal contribution.}\def\thefootnote{\arabic{footnote}}

\vspace{-10pt}
\section{Introduction}
\label{sec:intro}

\begin{figure}[t]
\begin{center}
\includegraphics[width=.99\linewidth]{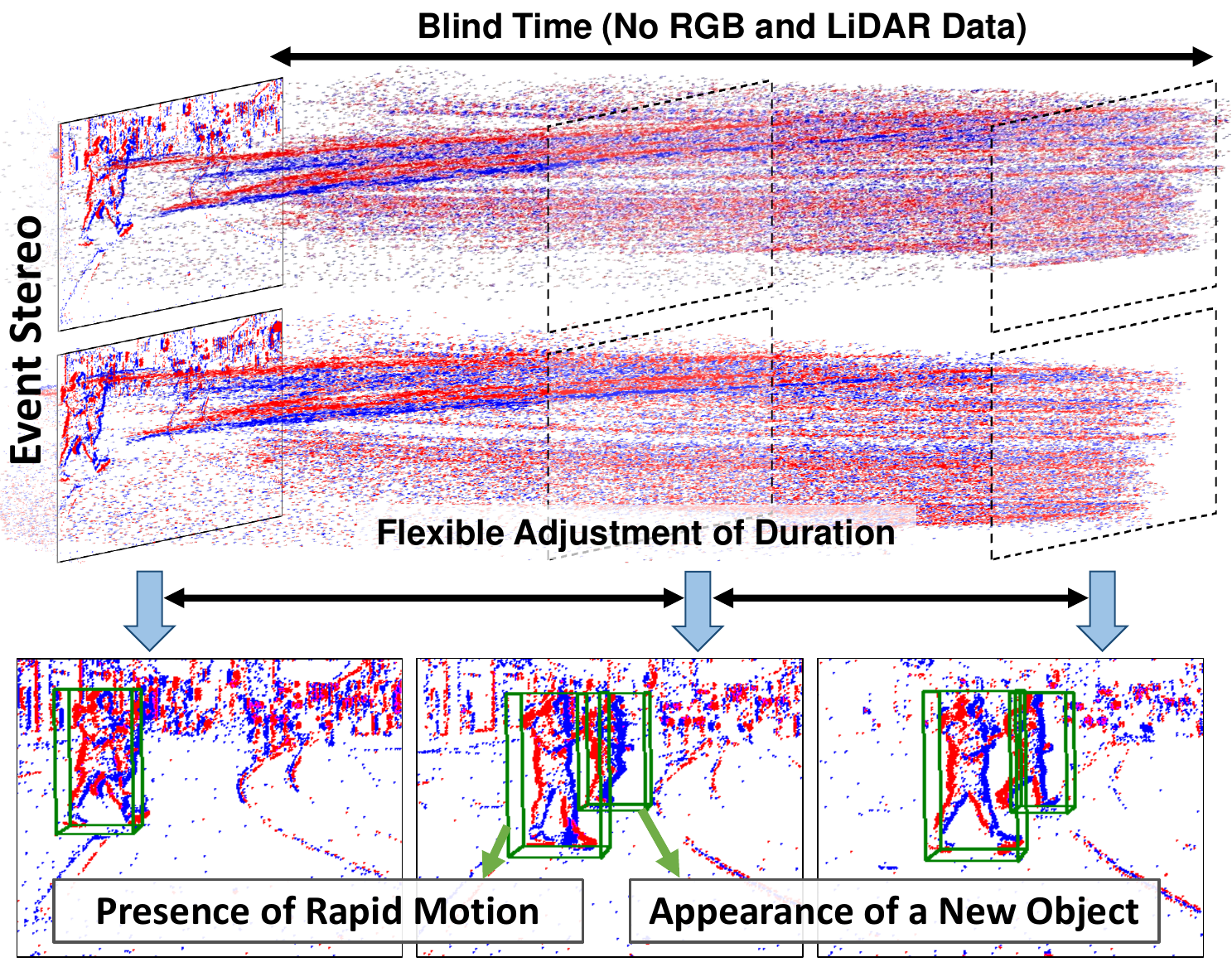}
\vspace{-5pt}
\caption{During blind time, our method enables fast and large motion-aware continuous 3D object detection at arbitrary points in time, including for newly appearing objects.}
\label{fig:teaser}
\end{center}
\vspace{-15pt}
\end{figure}

3D object detection is essential for accurately identifying the position and dimensions of surrounding objects, making it a fundamental task in autonomous systems.
Multi-modal sensors, such as LiDAR and RGB cameras, play a key role. However, fixed-frame rate sensors have discrete intervals during which data cannot be collected, resulting in inevitable detection delays.
This delay in capturing dynamic changes can significantly impact real-time decision-making, emphasizing the importance of high temporal resolution for reliable autonomous driving.

Event cameras~\cite{gallego2020event} are asynchronous sensors that continuously stream data over time, allowing them to break the trade-off between bandwidth and delay. Their high temporal resolution makes them well-suited for capturing fast motion, enabling high-speed perception even in dynamic scenarios. This capability is particularly beneficial for autonomous driving, where rapid response is critical for accident prevention and safe navigation. 

A recent work, Ev-3DOD~\cite{Cho_2025_CVPR}, integrates an RGB camera, LiDAR, and an event camera to overcome the frame-rate limitations of conventional sensors in 3D object detection. By incorporating event data, it enables perception even during periods when LiDAR and RGB cameras are unavailable. Notably, it introduces continuous-time 3D object detection, allowing inference at arbitrary timestamps before the next synchronized sensor data arrives. Continuous-time 3D object detection enables accurate inference beyond the timeframes when synchronized sensors, such as RGB and LiDAR, provide data. Ev-3DOD~\cite{Cho_2025_CVPR} defined periods with LiDAR and RGB data as active time and intervals without them as blind time, demonstrating that event cameras can facilitate 3D detection during these blind periods.

However, approaches that temporally propagate LiDAR- and image-based detection with events are highly dependent on synchronized sensors introducing limitations in dynamic environments.
For example, in scenarios like drones with rapid motion during adjacent active times, or high-speed driving causing significant movement during blind periods, fixed frame-rate sensor data does not fully reflect the current 3D geometry.
As a result, errors are accumulated over blind time, leading to frequent detection failures.

To this end, we propose an event stereo setup that performs precise 3D object detection using only spare and asynchronous event cameras to address the limitations of conventional sensors. 
As shown in Fig.~\ref{fig:teaser}, event stereo can compute 3D information even during blind time, enabling the detection of large-moving and newly appearing objects, which are scenarios that synchronized sensors cannot handle in blind time.

One of the challenges when using only event data for 3D object detection is its sparse nature. 
Event data lacks both semantic and geometric information compared to images and LiDAR, making accurate classification and regression particularly difficult. To overcome this, we propose a dual semantic-geometric filter module that collaboratively filters and enhances both information, fully leveraging stereo event data. The event semantic feature filters the geometric depth feature while the geometric feature simultaneously enhances the semantic information through the estimated disparity.
Additionally, following the global detection head, a dense, object-centric semantic event feature is employed to align the local 3D offsets of the bounding boxes, thereby enhancing regression performance.

Our proposed method achieves comparable performance in continuous-time 3D object detection without relying on rich fixed frame-rate sensor data. Specifically, through experiments in which the scene undergoes significant changes due to large motion within the same duration, as well as tests involving various intervals for inference, our approach outperforms existing methods. This result can be attributed to the use of asynchronous sensors, which offer distinct advantages in handling dynamic scenarios while fully utilizing both semantic and geometric information of events. The paper highlights the potential and scalability of event cameras to enhance 3D object detection in various environments.

\section{Related Works}
\label{sec:related_work}
\noindent
\textbf{Camera-based 3D Object Detection.}
Cameras offer significant cost advantages over LiDAR, leading many researchers to focus on developing methods for 3D object detection using image-only inputs~\cite{chen2022pseudo, lu2021geometry,  wang2022detr3d, liu2022petr, li2023bevdepth, huang2021bevdet, jiang2023polarformer, sun2020disp, wang2024bevspread, Liu_2024_CVPR}. However, image-based 3D detection faces challenges due to the lack of direct depth information. To address this, several works~\cite{chen2020dsgn, chen2022dsgn++, li2019stereo, liu2021yolostereo3d, guo2021liga} propose using stereo and multi-camera systems to compute 3D geometry through multiple views.
Despite these advancements, accurately estimating depth from images remains challenging, causing image-based methods to underperform compared to LiDAR-based approaches.

\begin{figure*}[t]
\begin{center}
\includegraphics[width=.99\linewidth]{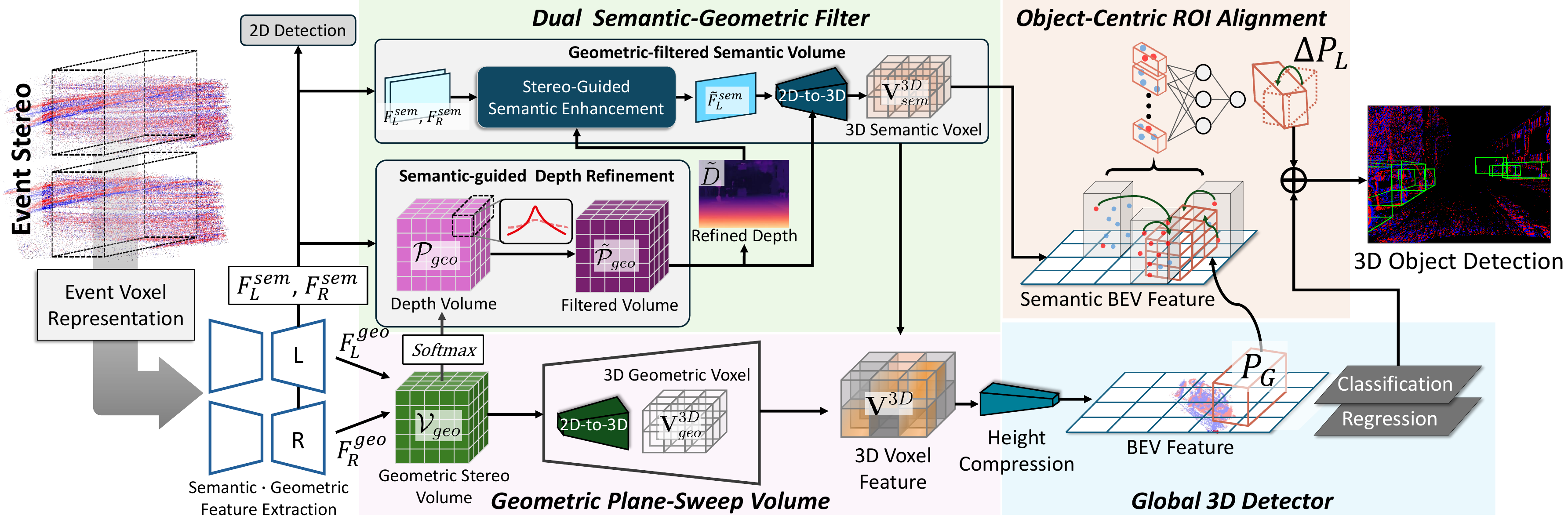}
\vspace{-3pt}
\caption{Overview of the proposed method. Our method is categorized into four components: Geometric Plane-Sweep Volume (Sec.~\ref{sec:geo_volume}), Dual Semantic-Geometric Filter (Sec.~\ref{sec:dual_filter}), Global 3D Detector (Sec.~\ref{sec:global_detect}), and Object-Centric ROI Alignment (Sec.~\ref{sec:roi_align}).}
\label{fig:overall_method}
\end{center}
\vspace{-15pt}
\end{figure*}

\noindent
\textbf{LiDAR-based 3D Object Detection.}
LiDAR-based 3D object detection can reliably estimate 3D bounding boxes using point clouds captured from LiDAR sensors. 

Research on representing 3D information from LiDAR has been conducted through various approaches, including point-based methods~\cite{shi2020point, shi2019pointrcnn, zhou2020joint, yin2022proposalcontrast, yang20203dssd, hu2022point, feng2024interpretable3d}, voxel-based methods~\cite{he2022voxel, lang2019pointpillars, li2024di, wang2023ssda3d, xu2022int, fan2022embracing, guan2022m3detr, he2022voxel, mao2021voxel, sheng2021improving, Zhang_2024_CVPR}, and point-voxel fusion networks~\cite{miao2021pvgnet, shi2020pv, yang2019std, he2020structure}.

To further boost the performance of LiDAR-based 3D detection, multi-modal sensor fusion~\cite{ huang2021makes, li2023logonet, Chen2022FUTR3DAU, Xie2023SparseFusionFM, li2024gafusion, yin2024fusion, lin2024rcbevdet, huang2025detecting, song2024graphbev} has emerged, where different sensors, such as cameras and LiDAR, complement each other. Fusion approaches are categorized into early~\cite{chen2022deformable, vora2020pointpainting, xu2021fusionpainting}, middle~\cite{li2022unifying, li2022voxel, li2022deepfusion, liang2022bevfusion, liu2023bevfusion, piergiovanni20214d, yang2022deepinteraction, prakash2021multi}, and late fusion~\cite{pang2020clocs, bai2022transfusion, li2023logonet}. Multi-modal 3D object detection has garnered significant attention for its ability to improve accuracy and enhance performance compared to unimodal methods.
However, there is still room for improvement in terms of safety in autonomous driving. LiDAR and cameras have limited time resolution (\ie, 10-20 Hz) because of their relatively high bandwidths. These limitations are especially noticeable in scenarios like high-speed driving or drone flight. To improve safety, algorithms that can operate at higher speeds are crucial.

\noindent
\textbf{Event-based Stereo.}  
Event cameras are well-suited for capturing dynamic scenes due to their high temporal resolution, low latency, and ability to operate effectively in challenging lighting conditions. Based on these characteristics, approaches using event cameras in a stereo setup~\cite{zhu2018realtime,zou2017robust, mostafavi2021event, nam2022stereo, ahmed2021deep, zhang2022discrete, cho2024temporal, cho2023learning, eomvs} have been proposed to better understand the surrounding 3D dynamic scenes. Additionally, research has explored using event cameras in a stereo setup alongside other sensors~\cite{zhao2024edge, cho2022selection, chen2024depth, lou2025zero, bartolomei2024lidar, cho2022event}, aiming to leverage the strengths of both event data and conventional sensors to enhance perception capabilities. Building on this research direction, we demonstrate that 3D object detection can be performed using only asynchronous event cameras. Furthermore, we highlight the potential of event cameras for 3D detection in dynamic and rapid motion scenarios.

\noindent
\textbf{Event-based Object Detection.} 
On the 2D image plane, numerous studies~\cite{ Gehrig2022PushingTL, Li2021GraphbasedAE, Li2022AsynchronousSM, Perot2020LearningTD, Iacono2018TowardsEO, Hamaguchi2023HierarchicalNM, Peng2023BetterAF, Wang2023DualMA, Zubic2023FromCC,  Gehrig2022RecurrentVT, Peng2023GETGE, Li2023SODFormerSO, cao2024embracing} have leveraged the event camera's sensitivity to moving objects for object detection. Various architectures, including graph-based~\cite{Schaefer22cvpr, Gehrig2024LowlatencyAV, Bi2019GraphBasedOC}, spiking~\cite{Yao2021TemporalwiseAS}, and sparse~\cite{Peng2024SceneAS} neural networks, have been explored to achieve streamlined and compelling results tailored to specific objectives. A recent study~\cite{Cho_2025_CVPR} demonstrated the advantages of high temporal resolution of event cameras in 3D object detection by integrating them with LiDAR and RGB cameras. This study addressed the limitation of synchronized sensors, such as RGB cameras and LiDAR, which operate at fixed frame rates and restrict inference to discrete time intervals. By integrating event cameras, the proposed approach overcomes this constraint, enabling continuous 3D detection even in the absence of synchronized sensor input. 

Despite these promising results, the approach remains partially dependent on synchronization, as critical information is solely derived from RGB and LiDAR. Consequently, significant errors may occur when rapid motion leads to substantial scene changes during blind time. To address this limitation, we propose the first fully asynchronous system for 3D object detection, utilizing a stereo event camera setup. Thanks to this asynchronous advantage, our method maintains strong performance even when the scene undergoes significant changes during the blind time or when inference is required at a finer continuous-time scale.

\section{Methods}
\label{sec:methods}

\subsection{Preliminaries}
\textbf{Event Camera.} Traditional frame-based cameras capture images at fixed intervals, whereas event cameras operate asynchronously, recording changes in logarithmic intensity \( L(u, v, t) \) only when they exceed a predefined threshold \( \mathcal{C} \). This can be expressed as:
\begin{equation}
L(u,v,t) - L(u,v,t-\Delta t) \geq \mathbf{p}\mathcal{C}, \quad \mathbf{p} \in \{-1,+1\},
\label{equ:event_prin}
\end{equation}
where \( \Delta t \) represents the time difference, and \( \mathbf{p} \) denotes event polarity, indicating either negative or positive events. The event stream \( \mathcal{E} \) consists of events \( e_i \) with spatial coordinates \((u_i, v_i)\), timestamp \( t_i \), and polarity \( p_i \):  
\begin{equation}
\mathcal{E} = \{e_i\}_{i=1}^N, \quad e_i = (u_i, v_i, t_i, p_i).
\label{equ:even_format}
\end{equation}
\textbf{Event Representation.} We adopt a voxel grid~\cite{zhu2019unsupervised} representation, where events are discretized along the time axis into \( B \) bins. Given \( N \) input events, timestamps are normalized to \([0, B-1]\), and the event volume is constructed as:
\begin{equation}
\quad E(u, v, b) = \sum _{i=1}^N p_i k_{\mathcal{B}}(u - u_i) k_{\mathcal{B}}(v - v_i) k_{\mathcal{B}}  (b - b^*_i),
\label{equ:voxel_grid}
\end{equation}
where \(b^*_i = (B - 1)(t_i - t_1)/(t_N - t_1)\) and \( k_{\mathcal{B}}(a) \) is a bilinear sampling kernel. The voxel grid allows efficient feature extraction via 2D convolution across spatial dimensions while conserving temporal information.

We use the event stream from the recent interval $\Delta \tau$, denoted as 
$\mathcal{E}_{(\tau - \Delta \tau) \rightarrow \tau} = \{{(u, v, t, p) \mid \tau - \Delta \tau \leq t < \tau}\}$
, to detect 3D boxes at time $\tau$. To achieve the continuous-time detection framework, \( \Delta \tau \) can adapt flexibly to different scenarios. This enables detection at arbitrary time points, making the framework adaptable to various situations. 
To simplify notation in this section, we represent the voxel grids generated from \( \mathcal{E}_{(\tau-\Delta \tau) \rightarrow \tau} \) as \( E_L \) and \( E_R \) for the left and right cameras, respectively.

\subsection{Geometric Plane-Sweep Volume}
\label{sec:geo_volume}
We illustrate the overall training pipeline in Fig.~\ref{fig:overall_method}. We modify the PSMNet~\cite{psmnet} to design our feature extractor and incorporated an additional feature head to disentangle the roles of semantics-based detection and geometric scene construction. 
This allows the extractor to separately extract semantic features $F_{L,R}^{sem}$ and geometric features $F_{L,R}^{geo}$, each with a dimension of $\mathbb{R}^{H \times W \times C}$.
The semantic feature activates well on objects through an auxiliary 2D dense head on the 2D image. The geometric feature activates across the entire scene on the stereo images, concentrating on finding correspondences.
First, we generate a geometric plane-sweep volume \(\mathcal{V}_{geo}\in\mathbb{R}^{H\times W\times D \times C}\) by concatenating left and corresponding right geometric features at each depth level \(D\), as done in previous works~\cite{chen2020dsgn, guo2021liga}. The volume is defined as:
\begin{equation}
\mathcal{V}_{geo}(u,v,w) = (F_L^{geo}(u,v) ||  F_R^{geo}(u - \frac{fL}{d(w)}, v)),
\label{equ:geo_volume}
\end{equation}
where \( (u, v) \) are pixel coordinates, $||$ denotes the concatenation, \( w = 0, 1, \dots \) is the candidate depth index, and \( d(w) = w \cdot v_d + z_{\text{min}} \) is the function to calculate the corresponding depth, where \( v_d \) is the depth interval and \( z_{\text{min}} \) is the minimum depth of the detection area. \( f \) and \( L \) represent the camera focal length and the baseline of the stereo camera pair, respectively.
From the geometric stereo volume \(\mathcal{V}_{\text{geo}}\), we apply 3D convolutions to reduce the channel dimension, followed by a softmax operation along the depth dimension, resulting in a depth probability volume \(\mathcal{P}_{\text{geo}} \in \mathbb{R}^{H \times W \times D}\), where \(\mathcal{P}_{\text{geo}}(u,v,:)\) represents the depth probability for pixel \((u, v)\) across all depth levels.

To transform the feature volume from stereo space to a 3D space, the 3D detection area is segmented into uniformly sized voxels, resulting in $\mathbf{V}^{3D}_{geo} \in \mathbb{R}^{X \times Y \times Z \times C}$. For each voxel, its center coordinates \((x, y, z)\) are projected back into the stereo space using the intrinsic parameters \(K\), resulting in the reprojected pixel coordinates \((u, v)\) and a corresponding depth index \(d^{-1}(z) = (z - z_{\text{min}})/v_d\) as: 
\begin{equation}
\mathbf{V}^{3D}_{geo}(x, y, z) = \mathcal{V}_{geo}(u, v, d^{-1}(z)).
\label{equ:geo_3d_volume}
\end{equation}

\begin{figure}[t!]
\begin{center}
\includegraphics[width=.99\linewidth]{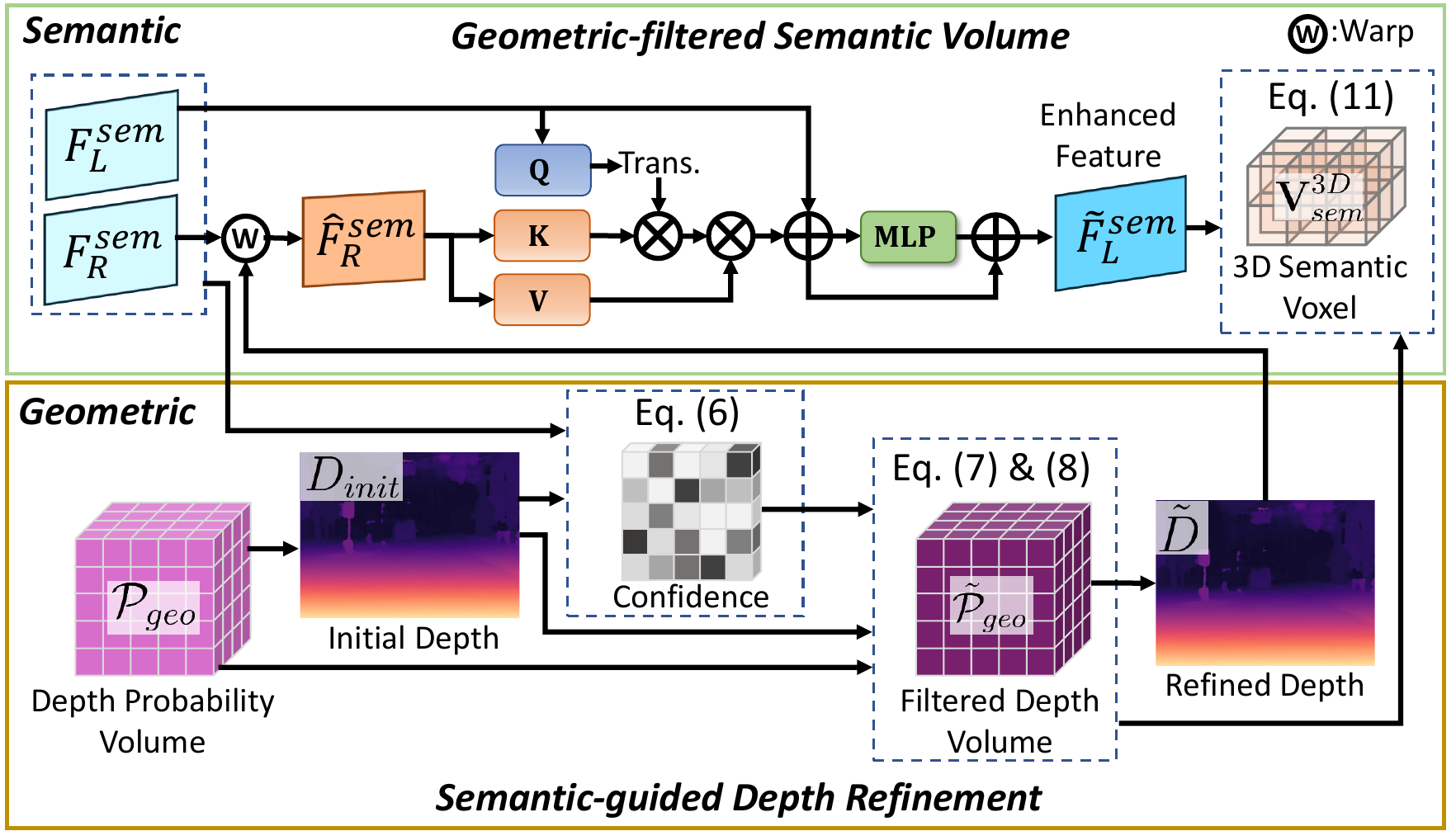}
\vspace{-5pt}
\caption{Dual semantic-geometric filter. Semantic and geometric information interact complementarily to enhance both the semantic features and the depth volume.}
\label{fig:dual_filter_method}
\end{center}
\vspace{-19pt}
\end{figure}

\subsection{Dual Semantic-Geometric Filter}
\label{sec:dual_filter}

Along with the geometric volume, we leverage the event camera's robustness in detecting objects on the 2D plane by using semantic features to construct a 3D semantic volume \( \mathbf{V}_{sem}^{3D}\in \mathbb{R}^{X \times Y \times Z \times C}\). Creating a 3D volume with only semantic features fails to capture geometric properties. To address this, we propose a dual semantic-geometric filter that combines both semantic and geometric information, as shown in Figure~\ref{fig:dual_filter_method}.

\noindent
\textbf{Semantic-guided Depth Refinement.}
One of the key challenges in 3D detection using stereo cameras is achieving accurate depth inference. Unlike LiDAR, stereo cameras do not provide direct depth measurements from the sensor. To enhance depth accuracy, we utilize semantic information to refine the geometric cost volume. As shown in 
Fig.~\ref{fig:dual_filter_method}, we first estimate the initial depth \( D_{init} \) by computing the weighted sum over all depth levels using the depth probability volume \( \mathcal{P}_{geo} \).
Additionally, to refine the depth distribution volume by transferring accuracy through the semantic feature, which effectively captures semantic information about the scene, we compute the similarity for each pixel based on the predicted depth as:
\begin{equation}
S(u,v) = \langle F_L^{sem}(u,v), F_R^{sem}(u - \frac{fL}{D_{init}(u,v)}, v) \rangle,
\label{equ:sem_sim}
\end{equation}
where \( \langle \cdot, \cdot \rangle \) denotes the inner product for similarity.
Similar to previous works~\cite{chen2023learning, shaked2017improved, poggi2021confidence,acvnet}
 that handle confidence in stereo matching, we also compute confidence as follows: 
\begin{equation}
C(u,v) = \sum_{w} \mathcal{P}_{geo}(u,v,w) \cdot \left( d(w) - D^{init}(u,v) \right)^2 ,
\label{equ:uncert}
\end{equation}
We aim to refine the depth distribution by considering both stereo similarity and confidence. Adapting the refinement process from~\cite{acvnet}, we generate filtered volume, $\tilde{\mathcal{P}}_{geo}$, by incorporating neighboring probabilities as follows:
\begin{equation}
\begin{aligned}
    & W_m(u,v) = S_m(u,v) \cdot \text{sigmoid} \left(C_m(u,v) \right), \\
    & \tilde{\mathcal{P}}_{geo}(u,v,w) = \sum_{m=1}^{M} \mathcal{P}_{geo}(u,v,w) \cdot \text{softmax}_m(W_m(u,v)),
\end{aligned}
\end{equation}
where $m=1,2,3, \dots, M$ are the sampled neighboring pixels and $\text{softmax}_m$ denotes a softmax across the $m$ dimension. 
Through this refinement process, depth ambiguity can be resolved, and we estimate the refined depth, \(\tilde{D}\), from the refined probability.

\noindent
\textbf{Geometric-filtered Semantic Volume.}
Although semantic features are well activated for objects, event data is sparse along the spatial dimension, often leading to ambiguity. To compensate for this lack of information, we enhance the left features using the estimated depth and the semantic features from the right camera. As shown in the bottom of Fig.~\ref{fig:dual_filter_method}, we warp the right semantic features to the left camera using the estimated depth: $\hat{F}_R^{sem} = \text{Warp}(F_R^{sem}, \tilde{D})$. However, directly using these warped features may introduce occlusion issues and fail to fully address potential misalignments between stereo cameras. To handle these challenges, we apply a transformer-based channel attention mechanism to generate enhanced semantic features. We generate the query, key, and value features as follows: \(\mathbf{Q}= W_Q(F_L^{sem})\), \(\mathbf{K}= W_K(\hat{F}_R^{sem})\), and \(\mathbf{V}= W_V(\hat{F}_R^{sem})\), where \(W_{(\cdot)}\) represents a \(1 \times 1\) convolution. The results of the transformer are computed in this manner:
\begin{align}
\mathbb{A}(F_L^{sem}, \hat{F}_R^{sem})= F_L^{sem} + W_\mathbb{A}(\operatorname{softmax} \left( \frac{\mathbf{Q}\mathbf{K}^{T}}{\alpha} \right) \cdot \mathbf{V})
\label{equ:attention_alignment}
\vspace{-2pt}
\end{align}
where \(\alpha\) is a learnable parameter. Finally, the geometrically enhanced semantic feature $\tilde{F}_L^{sem}$ is obtained:
\begin{align}
\tilde{F}_L^{sem} = F_L^{sem} + \operatorname{MLP}(\mathbb{A}) + \mathbb{A}.
\label{equ:final_attention}
\vspace{-2pt}
\end{align}
Similar to Eq.~(\ref{equ:geo_3d_volume}), voxels can be projected onto the image plane to map 2D semantic features to 3D. However, to ensure the transformation is based on geometric information, we apply a depth probability mask as:
\begin{equation}
\mathbf{V}^{3D}_{sem}(x, y, z) = \tilde{F}_L^{sem}(u,v)
\cdot \tilde{\mathcal{P}}_{geo}(u,v,d^{-1}(z))
\label{equ:sem_3d_volume}
\end{equation}

\begin{table*}[t]
\setlength{\aboverulesep}{-2.4pt}
\setlength{\tabcolsep}{9.3pt}
\setlength{\belowrulesep}{1pt}
\renewcommand{\arraystretch}{1.1}
\begin{center}
\caption{Comparison on the DSEC-3DOD dataset for vehicle (IoU = 0.7) and pedestrian (IoU = 0.5) detection on  3D and BEV. We follow the evaluation protocol of Ev-3DOD~\cite{Cho_2025_CVPR} for 100 FPS 3D detection in blind time. Mod. is an abbreviation for moderate, and easy and moderate represent difficulty levels.  VEH and PED represent vehicle and pedestrian, respectively.}
\vspace{-5pt}
\label{tab:main_dsec}
\resizebox{.99\linewidth}{!}{
\begin{tabular}{c|c|cc|cc|cc|cc}
\hline
\hline
\multirow{2}{*}{Modality} & \multirow{2}{*}{Methods} &   \multicolumn{2}{c|}{VEH $\text{AP}_{3D}$} & \multicolumn{2}{c|}{VEH $\text{AP}_{BEV}$} & \multicolumn{2}{c|}{PED $\text{AP}_{3D}$} & \multicolumn{2}{c}{PED $\text{AP}_{BEV}$}  \\
\cline{3-10}
 &   & Easy & Mod. & Easy & Mod. & Easy & Mod. & Easy & Mod. \\
\hline
\hline
\multirow{2}{*}{LiDAR} & VoxelNeXt~\cite{chen2023voxelnext} & 12.66	& 11.06 & 31.46	& 28.39 & 10.59	& 6.61 & 12.77	& 7.96  \\
& HEDNet~\cite{zhang2024hednet} & 14.31 & 12.96	& 29.94&	27.85& 10.16	&8.29&11.90	&6.89\\
\hline
\multirow{2}{*}{LiDAR+RGB} & Focals Conv~\cite{chen2022focal} & 12.71	&11.69	&26.14	&23.83&8.92&	5.50		&12.54&	7.83 \\
 & LoGoNet~\cite{li2023logonet} &  17.65	&15.39		&32.55&	29.19&	11.66&	8.20&15.09&	9.77\\
\hline
LiDAR+RGB+Event & Ev-3DOD~\cite{Cho_2025_CVPR} &  \textbf{29.53}	& \textbf{25.07}& \textbf{49.31}& \textbf{44.44} & \underline{18.42} & \underline{12.49} & \textbf{29.06} & \textbf{20.34} \\
\hline
\multirow{2}{*}{RGB-Stereo} & DSGN~\cite{chen2020dsgn} & 16.29	&13.45&	31.90&	27.53& 4.06	&2.42&	6.08&	3.74\\
& LIGA~\cite{guo2021liga} & 14.26	&10.98&		27.25&	25.11&	6.02&	3.58&8.73&	4.89\\
\hline
Event-Stereo & Ours & \underline{23.47}	& \underline{19.62} & \underline{40.13} & \underline{33.03} & \textbf{19.86} & \textbf{12.93} & \underline{22.91} & \underline{14.34} \\
\hline
\hline
\end{tabular}
}
\end{center}
\vspace{-15pt}
\end{table*}

\noindent
\textbf{Fusion of Semantic and Geometric Voxels.}
We generate two types of voxels in the 3D space: one specialized for geometric information, $\mathbf{V}^{3D}_{geo}$, and the other for semantic information, $\mathbf{V}^{3D}_{sem}$. To integrate the two voxels into a unified 3D voxel that effectively combines both aspects, we concatenated and then merged them through 3D convolution, resulting in the fused voxel, $\mathbf{V}^{3D}$.

\subsection{Global 3D Detector}
\label{sec:global_detect}
Inspired by recent stereo 3D detection approaches~\cite{chen2020dsgn, chen2022dsgn++, guo2021liga}, we designed an anchor-based detector. The bird's-eye view (BEV) feature strikes a balance between accuracy and computational efficiency in 3D detection for autonomous driving. The unified 3D voxel features $\mathbf{V}^{3D} \in \mathbb{R}^{X \times Y \times Z \times C}$ are downsampled along the height dimension to produce a BEV feature of size $(X, Z)$. 
We utilize fixed-size class anchors, determined based on the training set statistics, and assign a predefined number of anchors with varying sizes and orientations to each BEV feature location $(x,z)$.
Similar to previous works~\cite{guo2021liga, chen2022dsgn++, chen2020dsgn}, the dense head network performs a regression from the anchor $A = (x_a, y_a, z_a, h_a, w_a, l_a, \theta_a)$ to predict the offset $\Delta P_G = (\delta x, \delta y, \delta z, \delta h, \delta w, \delta l, \delta \theta)$. 
The final prediction is calculated by applying an offset to the anchor to determine the global position using the box decoding function \(g_d\), 
\begin{equation}
\begin{aligned}
 P_G  = g_d(A, & \Delta P_G)  = (x_a+\delta x, y_a + \delta y, z_a + \delta z, \\
  & h_ae^{\delta h}, w_ae^{\delta w}, l_a e^{\delta l}, \theta_a + \frac{\pi}{2} \tanh(\delta \theta) ).
\label{equ:det_func}
\end{aligned}
\end{equation}

\begin{figure}[t!]
\begin{center}
\includegraphics[width=.92\linewidth]{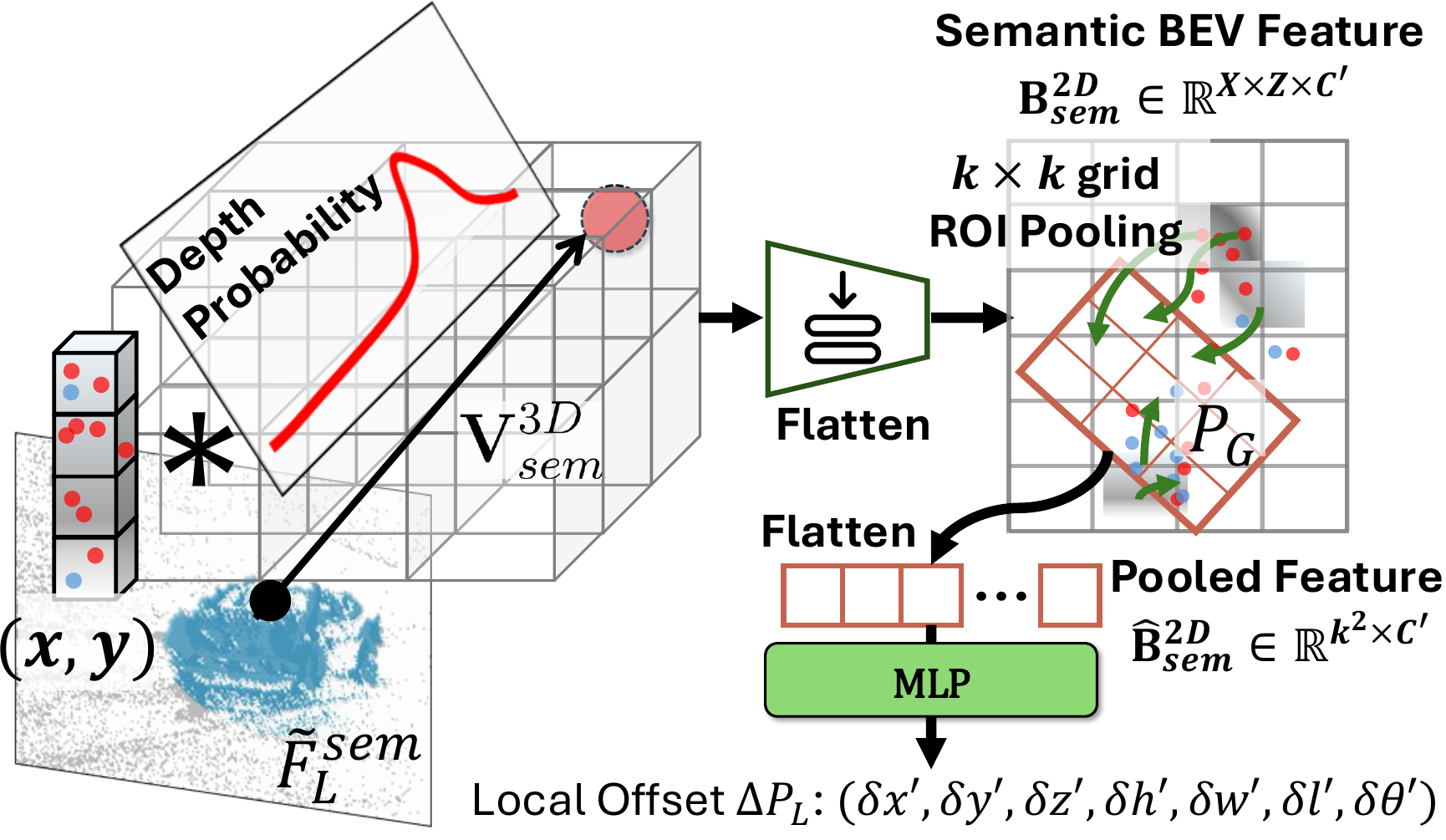}
\vspace{-5pt}
\caption{Illustration of the object-centric ROI alignment.}
\label{fig:roi_alignment_method}
\end{center}
\vspace{-19pt}
\end{figure}

\subsection{Object-Centric ROI Alignment}
 \label{sec:roi_align}
The geometric-semantic fused 3D voxels, $\mathbf{V}^{3D}$, generated through the semantic-geometric dual filter contain accurate depth and rich semantic information, enabling bounding box prediction in the global scene. 
However, a single-stage global regression can struggle with ambiguity caused by the motion of objects and the movement of the ego vehicle, especially as motion intensifies, necessitating a method that aligns more accurately with the object.
To this end, we propose a method for locally aligning the ROI around the object's bounding box.

\noindent
\textbf{ROI Pooling and Local Regression.}
As shown in Fig.~\ref{fig:roi_alignment_method}, we use the semantic voxel, $\mathbf{V}^{3D}_{sem}$, which focuses on the object and provides rich edge information from event-based semantic data, making it advantageous for ROI alignment. 
To optimize the computation of alignment, and because most stereo-based detections are prone to error along the $X$ and $Z$ dimensions, we project the voxel features onto the BEV, which corresponds to the transformation of $\mathbf{B}_{sem}^{2D}\in \mathbb{R}^{X \times Z \times C' }$ to the BEV of $\mathbf{V}_{sem}^{3D}$, rather than using them directly. Since $\mathbf{B}_{sem}^{2D}$ is splatted based on depth probabilities, it tends to cluster around the high-likelihood hypotheses of the predicted boxes. Therefore, to ensure that not only the exactly overlapping features but also the surrounding information is aggregated during alignment, we use a pooling scheme. Specifically, each bounding box prediction from global 3D detector, $P_G$, is split into an $k \times k$ voxel grid. The semantic BEV feature, $\mathbf{B}_{sem}^{2D}$, is then pooled~\cite{deng2021voxel} for each voxel grid. The pooled feature of each bounding box, $\hat{\mathbf{B}}_{sem}^{2D} \in \mathbb{R}^{k^2 \times C'}$, encodes object-centric local information, serving as a cue for precisely aligning the box. Each pooled feature passes through an MLP layer to predict the local alignment offset, $\Delta P_L = (\delta x', \delta y', \delta z', \delta h', \delta w', \delta l', \delta \theta')$. The final fine-grained regression, $\tilde{P}_G$, is calculated by box decoding function: $\tilde{P}_G = g_d(P_G, \Delta P_L)$

\subsection{Objective Function}
A multi-task loss was applied to train the network, which includes the geometric plane-sweep volume, the dual semantic-geometric branch, the global 3D detector head, and object-centric ROI alignment module as follows:
\begin{equation}  \mathcal{L}=\mathcal{L}_{depth}^{init}+\mathcal{L}_{depth}^{refine}+\mathcal{L}_{2D}+\mathcal{L}_{cls}+\mathcal{L}_{reg}^{global}+\mathcal{L}_{reg}^{local}.
\end{equation}
where $\mathcal{L}_{depth}^{init}, \mathcal{L}_{2D},  \mathcal{L}_{cls}$ are adapted from  uni-modal depth loss, 2D detection loss, and classification loss in~\cite{guo2021liga}.
For refined depth loss, we utilized $smooth\ L_1$ loss~\cite{kendall2017end} as
\begin{equation}
    \mathcal{L}_{depth}^{refine} =smooth_{L_1}(D_{gt}-\tilde{D}).
\end{equation}

The regression loss for the global 3D detector and object-centric ROI alignment are as follows:
\begin{equation}
    \mathcal{L}_{reg}^{global} =\mathcal{L}_{reg}(P_G, G),  \quad \mathcal{L}_{reg}^{local} =\mathcal{L}_{reg}(\tilde{P}_G, G) \ 
\label{equ:loss_global}
\end{equation}
where $G$ refers to 3D ground-truth bounding box and $\mathcal{L}_{reg}$ is 3D regression loss adapted from~\cite{guo2021liga}.
\section{Experiments}
\label{sec:experiments}

\subsection{Dataset}
\noindent
\textbf{DSEC-3DOD Dataset.}
The DSEC-3DOD dataset is the real-world event-based 3D detection dataset introduced in the Ev-3DOD~\cite{Cho_2025_CVPR}. 
It is based on the DSEC~\cite{gehrig2021dsec}, a stereo event dataset for driving scenarios. DSEC-3DOD contains manually created 10 FPS annotations for vehicle and pedestrian classes and provides blind-time annotations at 100 FPS through interpolation and refinement. We used the widely adopted KITTI metric~\cite{Geiger2012AreWR} from frame-based stereo methods, which measures performance in both BEV (bird-eye-view) and 3D for each class. 

\subsection{Experiment Setup}

\noindent
\textbf{3D Detection during Blind Time.}
Consistent with the training and evaluation protocol of existing event-based 3D object detection~\cite{Cho_2025_CVPR}, we train and evaluate using 100 FPS annotations during blind time, when no LiDAR data is available due to the fixed frame rate of LiDAR (\ie,~10 FPS). To achieve this, we sliced the event stream at \( \Delta \tau = 10 \) ms and converted it into a voxel grid.

\begin{table*}[t]
\setlength{\aboverulesep}{-2.5pt}
\setlength{\tabcolsep}{1.7pt}
\setlength{\belowrulesep}{2pt}
\renewcommand{\arraystretch}{1.3}
\begin{center}
\caption{Performance evaluation across various motion scales and time slices, presenting results for the easy difficulty level. Each entry corresponds to 3D / BEV detection results. VEH and PED represent vehicle and pedestrian, respectively.
}
\vspace{-5pt}
\label{tab:easy_motion}
\resizebox{.99\linewidth}{!}{
\begin{tabular}{c|c|c|cc|cc|c|cc|c}
\hline
\hline
Motion  & Time & \multirow{2}{*}{Class} &  \multicolumn{2}{c|}{LiDAR} & \multicolumn{2}{c|}{LIDAR+RGB} & \multicolumn{1}{c|}{LiDAR+RGB+Event} & \multicolumn{2}{c|}{RGB Stereo} & \multicolumn{1}{c}{Event Stereo}\\
\cline{4-11}
Scale &  Slice &  & VoxelNeXt~\cite{chen2023voxelnext} & HEDNet~\cite{zhang2024hednet} & Focals Conv~\cite{chen2022focal} & LoGoNet~\cite{li2023logonet}& Ev-3DOD~\cite{Cho_2025_CVPR} & DSGN~\cite{chen2020dsgn} & LIGA~\cite{guo2021liga} & Ours \\
\hline
\hline

\multirow{4}{*}{$\times 2$} & \multirow{2}{*}{$\times10$} & VEH & 5.05 / 13.00 & 6.03 / 12.87 & 5.95 / 12.48 & 5.97 / 13.57 & \underline{15.68} / \underline{30.59} & 7.33 / 17.36& 6.00 / 12.87 & \textbf{22.72} / \textbf{39.81}\\ 
&  & PED & 4.07 / 4.89 & 3.40 / 4.05 & 3.16 / 4.79 & 2.24 / 3.72 & \underline{7.88} / \underline{12.19} & 1.87 / 3.04 & 2.27 / 4.00\ & \textbf{19.04} / \textbf{22.71}\\
\cline{2-11}
 & \multirow{2}{*}{$\times20$} & VEH & 4.34 / 11.76 & 5.26 / 11.67 & 5.76 / 11.40 &5.42 / 12.40&\underline{15.96} / \underline{31.13}&7.13 / 16.43&5.67 / 12.00 & \textbf{23.47} / \textbf{40.13}\\ 
&  & PED & \underline{3.83} / \underline{4.31}&2.60 / 3.49&2.66 / 3.86&1.91 / 2.90&1.47 / 2.63&1.84 / 2.99&2.08 / 3.01& \textbf{19.86} / \textbf{22.91}\\
\hline
\multirow{4}{*}{$\times 4$} & \multirow{2}{*}{$\times10$} & VEH & 2.05 / 5.22&2.87 / 4.63&2.73 / 4.48&1.90 / 5.26&\underline{6.24} / \underline{11.62}&3.12 / 7.20&2.10 / 4.89 & \textbf{20.70} / \textbf{37.08}\\ 
&  & PED & 2.21 / 2.44&1.64 / 1.85&1.34 / 1.85&1.17 / 1.55&\underline{3.04} / \underline{4.09}&0.60 / 0.77&1.11 / 1.30 & \textbf{16.71} / \textbf{21.93}\\ 
\cline{2-11}
 & \multirow{2}{*}{$\times20$} & VEH & 1.71 / 4.55&2.27 / 4.29&2.73 / 3.94&1.90 / 4.66&\underline{6.36} / \underline{11.70}&3.09 / 6.52&1.54 / 4.17& \textbf{22.72} / \textbf{39.81} \\ 
&  & PED & \underline{2.14} / \underline{2.17}&1.10 / 1.42&1.02 / 1.44&0.53 / 1.28&0.39 / 0.43&0.59 / 0.75&0.82 / 1.12& \textbf{19.04} / \textbf{22.71}\\
\hline
\hline
\end{tabular}
}
\end{center}
\vspace{-10pt}
\end{table*}

\noindent
\textbf{Motion Scale and Time Slice.} 
The strength of event cameras lies in their ability to provide continuous information over time, even in dynamic motion~\cite{forrai2023event, bhattacharya2024monocular}. 
Therefore, experiments on dynamic and long-range motion during blind time are crucial to demonstrate the advantages of the asynchronous setup. However, the DSEC-3DOD dataset mostly lacks such scenes, as its average ego velocity is below 10 m/s. To address this, we introduce an experimental setup, \textbf{motion scale} (MS), which scales time axis for large motion experiments that closely resemble the real world. Accumulation and normalization of events over a longer period can approximate events with larger motion over the same time span. 
Therefore, increasing the motion scale in experiments can depict dynamic environments by skipping consecutive annotations and accumulating events over a longer period.
\textbf{Time slice} (TS) refers to the number of divisions applied to the blind time, where detection performance is evaluated at each time slice instance. Since various event input durations are used, this setup reflects the real-world demand for asynchronous detection. The baseline setup is defined with a motion scale of 1 and a time slice of 10, which corresponds to the 10 FPS fixed-frame rate data provided in the DSEC-3DOD dataset and the blind time annotations sliced into 10 parts (\ie, 100 FPS GT). 
To ensure compatibility with the original ground truth, experiments were conducted with motion scale values of 2 and 4 and time slice values of 10 and 20.
The model is trained using the original 100 FPS GT with a motion scale of 1 and a time slice of 10 and is evaluated across various motion scales and time slices. More details about the experimental settings are provided in the supplementary material.

\begin{figure}[t!]
\begin{center}
\includegraphics[width=.99\linewidth]{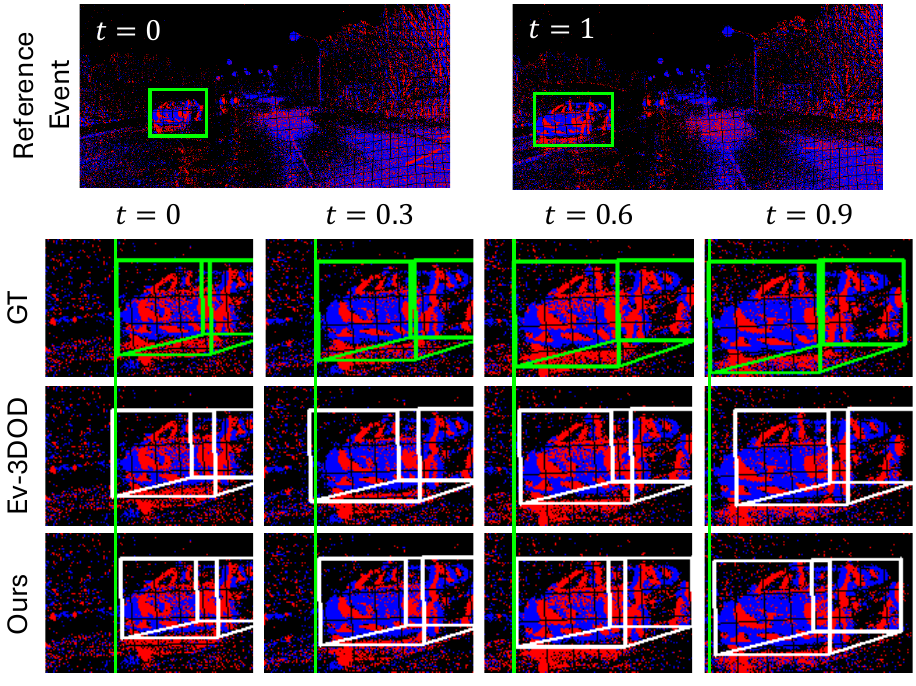}
\vspace{-6pt}
\caption{Comparison of 3D detection during blind time. $t = 0$ and $t = 1$ represent active time, while $t$ = 0.3, 0.6, and 0.9 indicate normalized blind time intervals.}
\label{fig:main_qual_motion1}
\end{center}
\vspace{-19pt}
\end{figure}

\subsection{Results on 3D Detection during Blind Time}
Table~\ref{tab:main_dsec} provides the results of 3D detection performed during blind time.
Results from previous works~\cite{chen2023voxelnext, zhang2024hednet, chen2022focal, li2023logonet} were taken from the supplementary material of Ev-3DOD~\cite{Cho_2025_CVPR}, while we trained and evaluated frame-based stereo methods, DSGN~\cite{chen2020dsgn} and LIGA~\cite{guo2021liga}, from scratch. Methods relying solely on synchronized sensors struggle to handle blind time, resulting in poor performance. 
On the other hand, both Ev-3DOD and our method utilize event cameras to overcome the limitations of fixed frame rates, enabling effective 3D detection during blind time and achieving high performance. While Ev-3DOD benefits from LiDAR-based depth information, generally outperforming our event-only approach, it is noteworthy that our method achieves superior performance in pedestrian AP\(_{3D}\). Pedestrian detection requires fine detail, and our high performance on this class is due to specialized modules that effectively leverage semantic features from event data.

Figure~\ref{fig:main_qual_motion1} illustrates another advantage of the asynchronous setup. While Ev-3DOD shows a decline in qualitative results as it moves further from active time 0, where LiDAR data is available, our method maintains high performance by computing geometric information from events, allowing it to operate independently of active time.

\begin{figure*}[t!]
\begin{center}
\includegraphics[width=.96\linewidth]{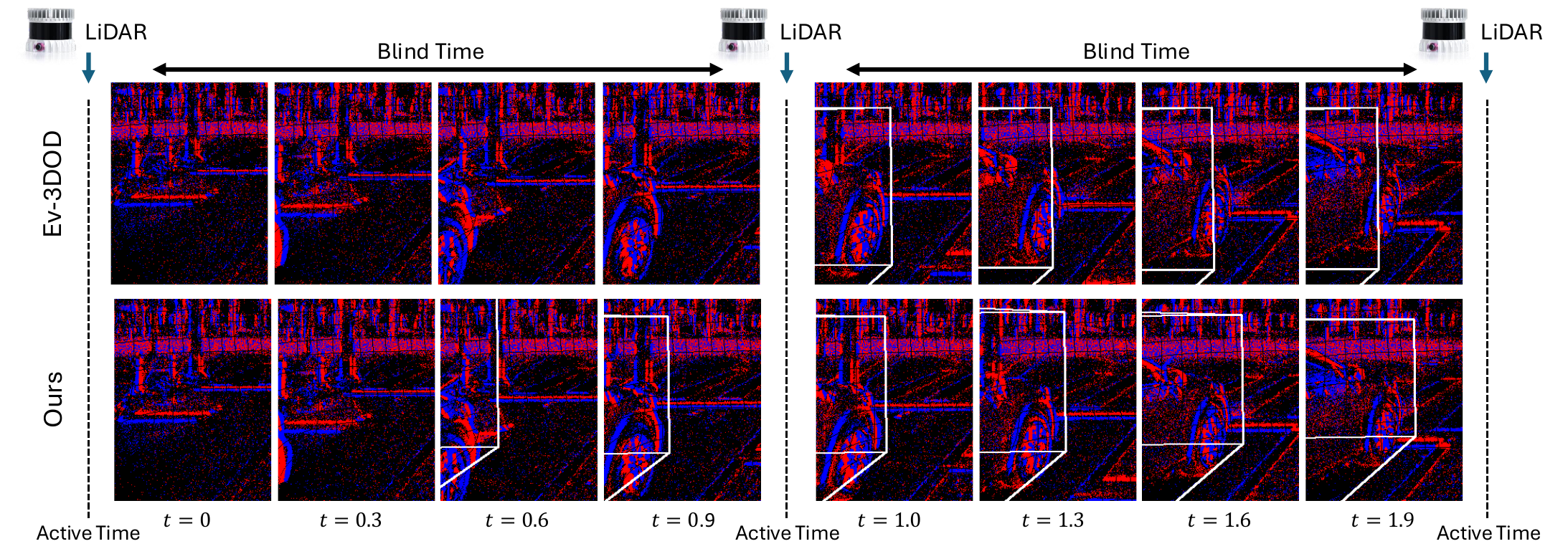}
\vspace{-6pt}
\caption{Comparison of 3D detection during blind time. Ev-3DOD detects new objects only at the active timestamps, $t = 0,1,2$, using LiDAR and RGB. In contrast, the proposed model can capture the emergence of new objects at arbitrary times.}
\label{fig:motion}
\end{center}
\vspace{-12pt}
\end{figure*}

\subsection{Diverse Motion Scales and Time Slices}
Table~\ref{tab:easy_motion} shows the comparison of ours with different methods across various motion scales and time slices. Methods relying solely on synchronized sensors~\cite{chen2023voxelnext, zhang2024hednet, chen2022focal, li2023logonet, chen2020dsgn, guo2021liga} exhibit a significant performance drop as the motion scale increases during blind time. 
Our primary comparison method, Ev-3DOD~\cite{Cho_2025_CVPR}, utilizes event data during blind time, allowing it to handle moderate scenarios well, as shown in Table~\ref{tab:main_dsec}. However, as the motion increases, its performance significantly drops. This is because Ev-3DOD relies entirely on LiDAR for 3D geometric information, and when large motion occurs, scene changes make past geometric information less useful for the present. In contrast, our method computes 3D geometric information using asynchronous event data, resulting in more stable performance across different motion scales. Therefore, in scenarios with high-speed driving or significant scene changes due to dynamic motion, our event-based 3D detection proves highly effective during blind time.

\section{Ablation Study}
\label{sec:ablation_study}

\begin{table}[t]
\begin{center}

\caption{Ablation study on modules for 3D detection during blind time. DSGF: Dual Semantic-Geometric Filter. SDR: Semantic-Guided Depth Refinement. GSV: Geometric-Filtered Semantic Volume. OCRA: Object-Centric ROI Alignment. The fused voxel source is denoted as G or S, where G represents geometric voxel and S represents semantic voxel.}
\label{tab:able_components}
\vspace{-6pt}
\renewcommand{\tabcolsep}{10.5pt}
\renewcommand{\arraystretch}{1.1}
\resizebox{.99\linewidth}{!}{
\begin{tabular}{cc|c|c|cc}
\Xhline{2.5\arrayrulewidth}
\multicolumn{2}{c|}{DSGF} & 3D Voxel& \multirow{2}{*}{OCRA} & \multicolumn{2}{c}{mAP Easy}  \\
\cline{1-2} 
SDR & GSV & Source & & 3D & BEV\\
\hline 
 & & G & &12.92	& 20.78\\ 
   &  & G+S & & 14.60 & 24.85 \\
   &  & G & \checkmark & 14.56	& 21.73\\ 
  & & G+S & \checkmark & 15.95 & 26.12\\ 
\checkmark & & G+S & & 15.85 & 27.78\\
 & \checkmark & G+S & & 17.08 & 26.53 \\
\checkmark & \checkmark & G+S & & \underline{19.12} & \underline{29.64} \\
\checkmark & \checkmark & G+S & \checkmark & \textbf{21.66} & \textbf{31.52} \\
\Xhline{2.5\arrayrulewidth}
\end{tabular}
}
\end{center}
\vspace{-10pt}
\end{table}

\noindent
\textbf{Ablation Study on Modules.}
As shown in Table~\ref{tab:able_components}, we conduct an ablation study on the proposed modules. Starting from a baseline that performs 3D detection using only the geometric plane-sweep volume, we incrementally add the proposed modules to evaluate their impact on performance. 

We report the easy mAP performance, which is the average value across the two classes, VEH and PED, measured under a motion scale of $1$ and a time slice of $10$. 
Compared to the baseline, simply adding semantic information results in sub-optimal performance, yielding only a 1.68 improvement in 3D detection. However, incorporating a dual semantic-geometric filter, which facilitates interaction between semantic and geometric features, leads to a significant performance gain of 6.2 in 3D detection. Additionally, the proposed object-centric ROI alignment consistently improves performance across all settings. When used alongside a dual semantic-geometric filter, it further enhances semantic BEV features, resulting in a more significant performance boost.

\begin{table}[t]
\renewcommand{\arraystretch}{1.05}
\begin{center}
\caption{Performance of RGB fusion. Results of easy difficulty level with a time slice of 10.}
\vspace{-2pt}
\label{tab:rgb_fusion}
\resizebox{.97\linewidth}{!}{
\begin{tabular}{c|c|cc|cc}
\hline
\multirow{2}{*}{Settings} & \multirow{2}{*}{Methods} & \multicolumn{2}{c|}{$\text{AP}_{3D}$} & \multicolumn{2}{c}{$\text{AP}_{BEV}$} \\
& & VEH & PED & VEH & PED \\
\hline
\multirow{2}{*}{Motion $\times 1$}  & Ours (E) & 23.47 & 19.86 & 40.13 & 22.91 \\
 & Ours (E+I)& \textbf{25.23} & \textbf{21.50} & \textbf{46.21} & \textbf{29.14}\\
\hline
\hline
\multirow{2}{*}{Motion $\times 2$} & Ours (E) & \textbf{22.72} & \textbf{19.04} & \textbf{39.81} & \textbf{22.71} \\
 & Ours (E+I)& 20.14  & 10.50 & 37.93 & 12.76\\
\hline
\end{tabular}
}
\end{center}
\vspace{-20pt}
\end{table}

\noindent
\textbf{RGB Fusion.} 
Table~\ref{tab:rgb_fusion} presents the results of fusing synchronous RGB information into our method. The stereo RGB features are extracted using a separate feature extractor and concatenated with the event features. This fusion provides improved performance under small motion scales, as the RGB input offers additional semantic guidance. However, under large motion, the performance significantly degrades due to misalignment between the synchronous RGB data and the actual object location. In contrast, our setup that relies solely on event data demonstrates robust performance regardless of scene dynamics.

\begin{figure}[t]
\begin{center}
\vspace{0pt}
\includegraphics[width=.87\linewidth]{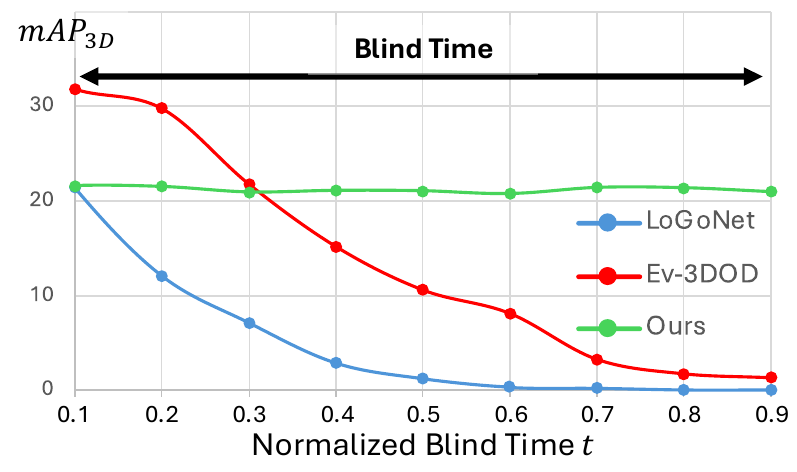}
\vspace{-8pt}
\caption{Performance comparison with other methods during blind time under the setting of motion scale 2 and time slice 10.}
\label{fig:frame_ablation}
\end{center}
\vspace{-10pt}
\end{figure}

\noindent
\textbf{Performance Degradation over Blind Time.} 
Figure~\ref{fig:frame_ablation} shows the performance over time during blind time for our method and other methods~\cite{Cho_2025_CVPR, li2023logonet}. 
To highlight robustness in challenging scenarios, we use motion scale of \(2\). $t = 0$ refers to the most recent synchronized sensor input timestamp, while $t = 0.1\sim0.9$ represents the blind interval before the next input.
The reason for the performance drop of other methods is that they rely on 3D information from synchronous sensors and struggle as the current scene diverges from recent active time ($t = 0$), making it challenging to estimate the present 3D box using past information. Thus, despite using event data, Ev-3DOD struggles to transfer past 3D information to the present, causing its performance to resemble LoGoNet as motion increases over time. In contrast, our method asynchronously infers both geometric and semantic information for 3D detection, allowing it to operate independently of active time.

\section{Conclusion}
\label{sec:conclusion}

In this paper, we tackle 3D object detection using event stereo in a fully asynchronous setting for the first time. Unlike the previous continuous-time 3D object detection work that utilizes LiDAR, our approach does not rely on LiDAR, leading to lower performance compared to methods that fuse LiDAR and event data. However, by evaluating various motion scales and time slices, we demonstrate that our method is more adaptable and performs robustly in diverse and extreme conditions. Furthermore, we highlight the potential of event-based 3D perception in dynamic environments where event cameras are particularly beneficial, such as drone flights, high-speed driving, and robots. We hope this work inspires further research in this direction.

\section{Acknowledgments. }
This work was supported by the National Research Foundation of Korea(NRF) grant funded by the Korea government(MSIT) (NRF2022R1A2B5B03002636). This work was also supported by the Technology Innovation Program (1415187329,20024355, Development of autonomous driving connectivity technology based on sensor-infrastructure cooperation) funded By the Ministry of Trade, Industry \& Energy(MOTIE, Korea).

{
    \small
    \bibliographystyle{ieeenat_fullname}
    \bibliography{main}
}

\maketitlesupplementary

\setcounter{section}{0}
\setcounter{figure}{0}
\setcounter{equation}{0}
\setcounter{table}{0}

\section{Details of Motion Scale and Time Slice}

\begin{figure*}[t]
\begin{center}
\includegraphics[width=.80\linewidth]{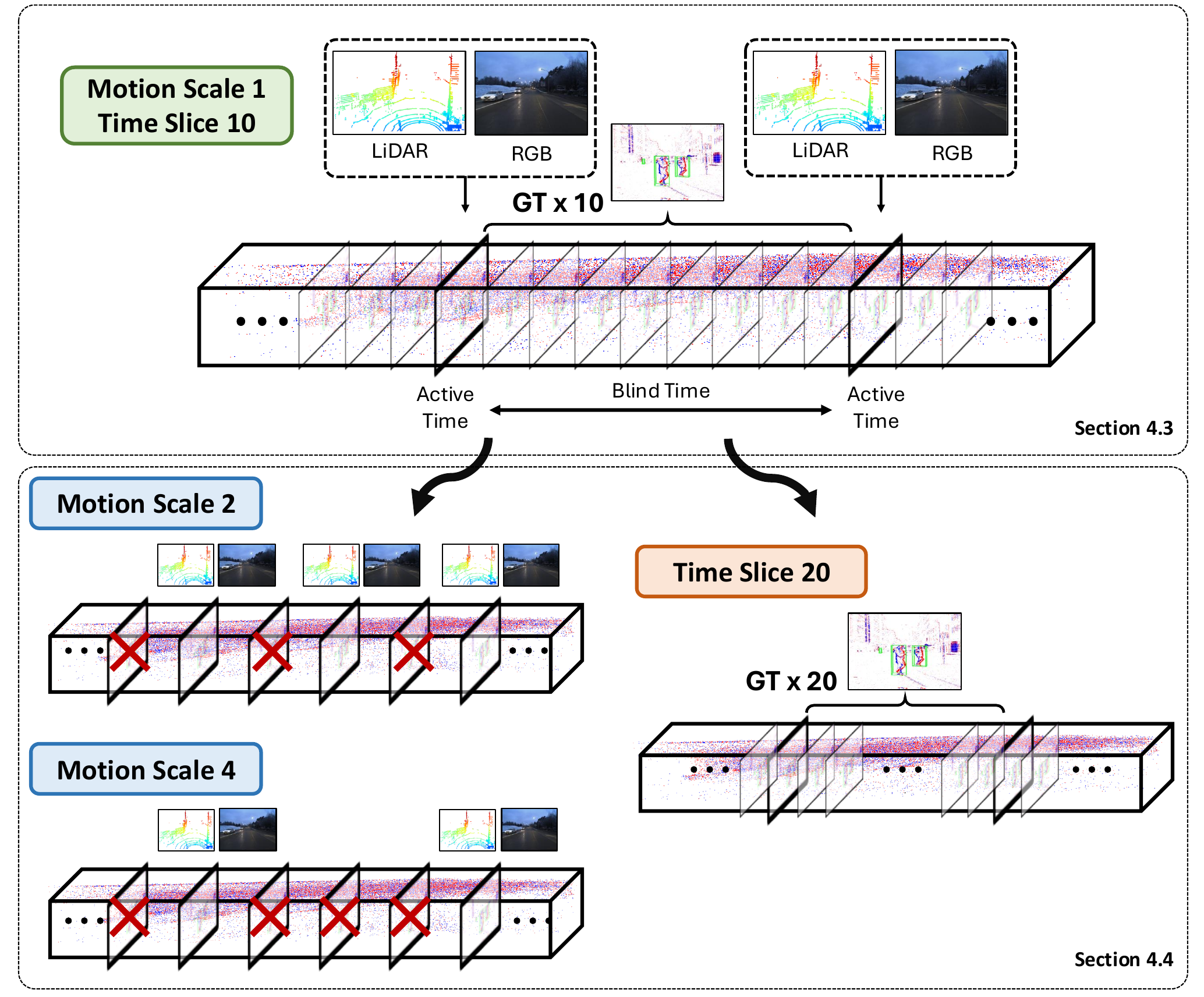}
\vspace{-6pt}
\caption{Visualization of the motion scale and time slice used in the experimental setup. The upper part of the figure represents the baseline setup following Ev-3DOD~\cite{Cho_2025_CVPR}, with a motion scale of 1 and a time slice of 10. The lower part illustrates the setup adopted to evaluate the model under large motion and diverse event inputs.}
\label{fig:supple_motion_time_slice}
\end{center}
\vspace{-10pt}
\end{figure*}

Figure~\ref{fig:supple_motion_time_slice} provides details of our experimental setup. To ensure a fair comparison with previous works, Sec.
~\textcolor{iccvblue}{4.3} of the main paper adopts the same evaluation protocol. DSEC-3DOD provides a fixed-frame-rate sensor at 10 FPS and blind time annotations at 100 FPS. Provided LiDAR and RGB data are fully utilized, and each blind time is evaluated with 10 ground truth annotations.

The evaluation setup of Ev-3DOD~\cite{Cho_2025_CVPR} allows for assessing detection performance during blind time. However, due to the limited motion in DSEC data and its fixed time intervals, model performance can only be evaluated at restricted points in time. To enable evaluation under asynchronous and diverse temporal conditions, we define motion scale and time slice as key evaluation setup parameters.

Motion scale is a parameter that controls scene motion by adjusting the length of the blind time. This is achieved by skipping consecutive frames of LiDAR and RGB data, thereby modifying the amount of motion. Events are accumulated over the blind time and normalized in the temporal domain. 
Motion scale control~\cite{cho2023learning, hagenaars2021self, zhu2019unsupervised, tulyakov2021time, shiba2022secrets, zhu2018ev}, which accumulates data over a longer period to represent large motion, is a well-established method widely used in other works for evaluating performance under large motion conditions. Therefore, following previous work, we also adopted motion scale control to represent dynamic and long-range motion.

Time slice controls the evaluation interval within the blind time. Each blind time is evaluated at multiple points determined by the time slice. This parameter introduces variations in the distribution of event data, making it a challenging factor for assessing the temporal flexibility of event-based methods.

We define the baseline experimental setup with a motion scale of 1 and a time slice of 10. Evaluations were conducted using various experimental parameters within the constraints of the given fixed-frame-rate sensor and available annotations. To ensure a fair evaluation, we assessed detection performance using only the model trained on the baseline setup.

\section{Implementation Details}

\textbf{Training Details.} 
Training was conducted on two NVIDIA TITAN RTX GPUs for 60 epochs with a batch size of 2. The AdamW optimizer~\cite{loshchilov2017decoupled} was employed with a learning rate set to 0.001. 

\noindent
\textbf{Depth Refinement.} 
To perform actual refinement, we computed a finer probability by considering neighboring probabilities. Thus, the practical implementation of Eq. (\textcolor{iccvblue}{6}) in the main paper incorporates neighboring pixels as follows:
\begin{equation}
S(u,v) =  \langle F_L^{sem}(u,v), F_R^{sem}(u - \frac{fL}{D_{init}^m(u,v)}, v) \rangle,
\label{equ:sem_sim}
\end{equation}
We use the $m = 1, 2, 3, 4, 5$ for neighboring sampling. 

\noindent
\textbf{Event Grid size and Voxel Size.} Following previous works~\cite{Cho_2025_CVPR}, we use the bin size of event voxel grid as 5. 3D geometric voxel and 3D semantic voxel in Sec.~\textcolor{iccvblue}{3.3} have range of $[-30.4m, 30.4m]$ in X axis, $[-1.0m, 3.0m]$ in Y axis, and $[2.0m, 56.9m]$ for Z axis. Voxel size is set to $(0.2m, 0.2m, 0.2m)$.

\noindent
\textbf{ROI Pooling for Alignment}
In Sec.~\textcolor{iccvblue}{3.5} of the main paper, the ROI \( P_G \) estimated by the global detector is divided into a \( k \times k \) voxel grid, for \( k=3 \). The semantic BEV features are pooled for each grid, and all grid features are aggregated to estimate the local offset.

\noindent
\textbf{Anchor Size}
As mentioned in the main paper, we use anchors with fixed size, height and orientation for each \((x, z)\) coordinate in the 3D voxel space. The fixed anchor sizes are determined by computing the class-wise box statistics from the training set. Anchors for vehicle class and pedestrian class are as follows:
\begin{equation}
\begin{aligned}
    &A_{veh}=(x, 0.47, z, 1.79, 1.86, 4.28, \{0, \frac{\pi}{2}\}) \\
    &A_{ped}=(x, 0.6, z, 1.73, 0.6, 0.8, \{0, \frac{\pi}{2}\})
\end{aligned}
\end{equation}


\section{Effectiveness of Semantic-guided Depth Refinement}
Following the KITTI stereo metric~\cite{Geiger2012AreWR}, we measure a depth estimation accuracy if depth error is below a specified outlier threshold. Table~\ref{tab:supple_depth_refine} compares performance across different outlier thresholds, highlighting the impact of semantic-guided depth refinement (SDR). The results show that SDR consistently enhances accuracy across all thresholds. Moreover, the depth refinement module improves not only the final detection performance but also the overall depth estimation quality.



\begin{table}[h!]
\begin{center}
\caption{Effectiveness of semantic-guided depth refinement. SDR: Semantic-Guided Depth Refinement.}
\label{tab:supple_depth_refine}
\vspace{-6pt}
\renewcommand{\tabcolsep}{9.5pt}
\renewcommand{\arraystretch}{1.1}
\resizebox{.99\linewidth}{!}{
\begin{tabular}{c|cccc}
\Xhline{2.5\arrayrulewidth}
Outlier Threshold & $>1.6m$ & $>0.8m$ & $>0.4m$ & $>0.2m$  \\
\hline 
 w/o SDR & 0.236 & 0.382 & 0.549 & 0.715\\
\textbf{ w/ SDR (Ours)} & \textbf{0.219} & \textbf{0.361} & \textbf{0.529} & \textbf{0.696} \\
\hline
\end{tabular}
}
\end{center}
\vspace{-20pt}
\end{table}

\section{Visualization on Semantic and Geometric Features}

\begin{figure*}[t!]
\begin{center}
\includegraphics[width=.99\linewidth]{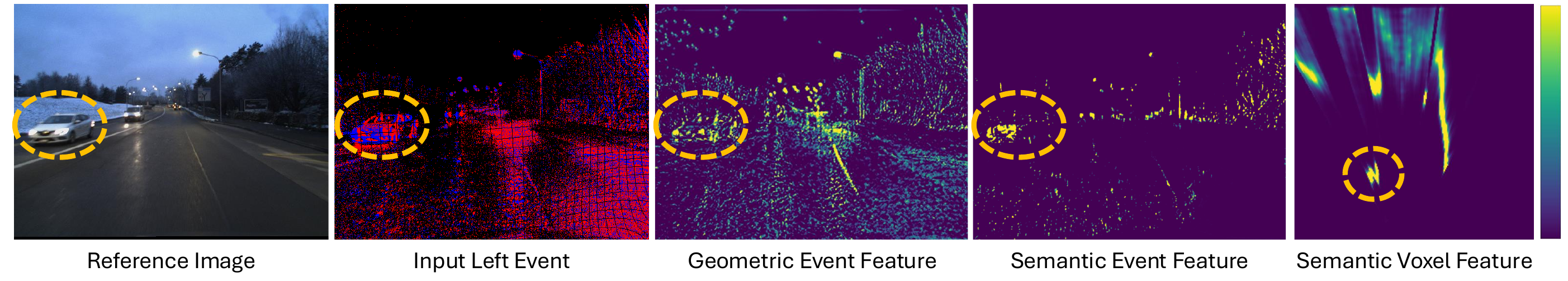}
\vspace{-6pt}
\caption{Example of input event, geometric feature, enhanced semantic feature, and semantic BEV feature. Feature values are normalized to \([0, 1]\) for visualization.}
\label{fig:supple_feature}
\end{center}
\vspace{-10pt}
\end{figure*}

Semantic and geometric event features serve distinct roles, and the model utilizes them collaboratively to enhance both geometric and semantic information. To provide insights into the characteristics of these features, we present visualizations of the extracted representations. Geometric features highlight complex structures that aid in stereo matching, whereas semantic features exhibit strong attention to target objects. By leveraging such object-centric information, ROI alignment is achieved, enabling fine-grained box regression.

\begin{table*}[t]
\setlength{\aboverulesep}{-2.5pt}
\setlength{\tabcolsep}{1.7pt}
\setlength{\belowrulesep}{2pt}
\renewcommand{\arraystretch}{1.3}
\begin{center}
\caption{Performance evaluation across various motion scales and time slices, presenting results for the moderate difficulty level. Each entry corresponds to 3D / BEV detection results. VEH and PED represent vehicle and pedestrian, respectively.}
\vspace{-5pt}
\label{tab:moderate_motion}
\resizebox{.99\linewidth}{!}{
\begin{tabular}{c|c|c|cc|cc|c|cc|c}
\hline
\hline
Motion  & Time & \multirow{2}{*}{Class} &  \multicolumn{2}{c|}{LiDAR} & \multicolumn{2}{c|}{LIDAR+RGB} & \multicolumn{1}{c|}{LiDAR+RGB+Event} & \multicolumn{2}{c|}{RGB Stereo} & \multicolumn{1}{c}{Event Stereo}\\
\cline{4-11}
Scale &  Slice &  & VoxelNeXt~\cite{chen2023voxelnext} & HEDNet~\cite{zhang2024hednet} & Focals Conv~\cite{chen2022focal} & LoGoNet~\cite{li2023logonet}& Ev-3DOD~\cite{Cho_2025_CVPR} & DSGN~\cite{chen2020dsgn} & LIGA~\cite{guo2021liga} & Ours \\
\hline
\hline

\multirow{4}{*}{$\times 2$} & \multirow{2}{*}{$\times10$} & VEH & 4.28 / 12.58&5.24 / 12.06&5.60 / 11.66&4.93 / 12.21&\underline{13.52} / \underline{26.56}&7.02 / 16.03&5.52 / 11.93 & \textbf{19.31} / \textbf{32.47}\\ 
&  & PED & 2.43 / 3.13&1.70 / 2.48&2.14 / 2.81&1.71 / 2.62&\underline{4.91} / \underline{8.57}&1.27 / 1.88&1.75 / 2.29 & \textbf{12.56} / \textbf{13.99}\\ 
\cline{2-11}
 & \multirow{2}{*}{$\times20$} & VEH &3.80 / 11.54&4.67 / 11.08&5.56 / 10.79&4.60 / 11.23&\underline{14.50} / \underline{27.93}&6.75 / 14.36&4.77 / 11.08 & \textbf{19.62} / \textbf{33.03}\\ 
&  & PED & \underline{2.31} / 2.48&1.49 / 2.18&1.52 / 2.15&1.44 / 2.24&1.62 / \underline{3.19}&1.26 / 1.77&1.42 / 2.12& \textbf{12.93} / \textbf{14.34}\\
\hline
\multirow{4}{*}{$\times 4$} & \multirow{2}{*}{$\times10$} & VEH & 2.03 / 5.22&2.85 / 4.78&2.73 / 3.96&1.88 / 4.37&\underline{5.59} / \underline{10.50}&3.08 / 6.27&1.82 / 4.50 & \textbf{16.42} / \textbf{29.01}\\ 
&  & PED & 0.91 / 1.18&0.91 / 0.91&0.91 / 1.06&0.91 / 0.91&\underline{2.31} / \underline{3.37}&0.41 / 0.49&0.91 / 1.05 & \textbf{10.61} / \textbf{13.86}\\ 
\cline{2-11}
 & \multirow{2}{*}{$\times20$} & VEH & 1.59 / 4.29&2.27 / 4.09&2.73 / 3.55&1.88 / 3.99 & \underline{5.60} / \underline{10.26}&3.06 / 5.62&1.36 / 3.87& \textbf{19.31} / \textbf{32.47}\\ 
&  & PED & \underline{0.91} / 0.91&\underline{0.91} / 0.91&\underline{0.91} / 0.91&0.45 / 0.91&\underline{0.91} / \underline{1.31}&0.34 / 0.46&0.45 / 0.69& \textbf{12.93} / \textbf{14.34}\\
\hline
\hline
\end{tabular}
}
\end{center}
\vspace{-10pt}
\end{table*}

\section{Additional Results}

\textbf{Quantitative Results. }
Table~\ref{tab:moderate_motion} presents the quantitative results of DSEC-3DOD at the moderate difficulty level, evaluated under various motion scales and time slices. Compared to our method, other approaches exhibit more significant performance degradation under large motions and longer blind times, as they heavily rely on synchronized sensors (\eg, RGB and LiDAR).

\noindent
\textbf{Qualitative Results. }

\begin{figure*}[t!]
\begin{center}
\includegraphics[width=.99\linewidth]{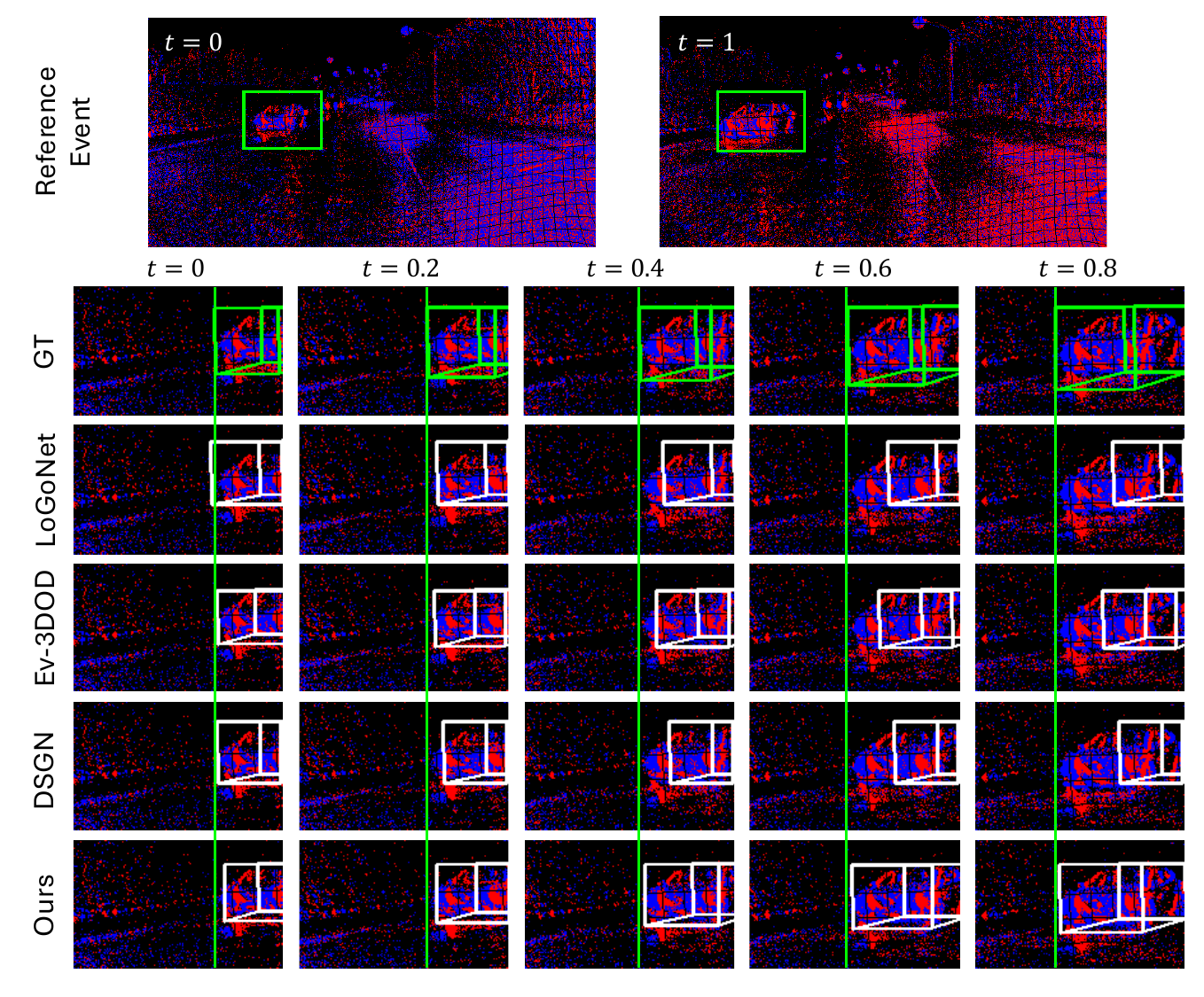}
\vspace{-6pt}
\caption{Comparison of 3D detection during blind time. The motion scale and time slice are set to 2 and 10, respectively. Green vertical lines across the image were added to compare the box's relative position. Fixed-frame-rate sensor-based methods (\ie, LoGoNet~\cite{li2023logonet} and DSGN~\cite{chen2020dsgn}) fail to predict objects during the blind time. Ev-3DOD~\cite{Cho_2025_CVPR} leverages monocular event data to propagate detection through blind time, but its performance deteriorates under large movements. The proposed method operates in a fully asynchronous manner, consistently producing stable results regardless of the blind time.}
\label{fig:supple_motion2}
\end{center}
\vspace{-10pt}
\end{figure*}

We provide additional qualitative results for the motion scale 2 and time slice 10 setup. The results demonstrate that conventional sensor-based methods suffer from significant detection errors due to large motion. Furthermore, compared to Fig.~\textcolor{iccvblue}{5} in the main paper, where the motion scale is set to 10, Ev-3DOD exhibits a substantial performance drop despite utilizing event data, as it remains heavily dependent on LiDAR. In contrast, our fully asynchronous model seamlessly adapts to large motion, ensuring robust detection.

{
    \small
    \bibliographystyle{ieeenat_fullname}
}

\end{document}